\documentclass[conference]{IEEEtran}

\IEEEoverridecommandlockouts

\usepackage[absolute,showboxes]{textpos}
\TPMargin{5pt}
\newcommand{\copyrightstatement}{
\begin{textblock}{0.84}(0.08,0.93)    
\noindent
\footnotesize
\copyright 2020 IEEE. Personal use of this material is permitted. Permission from IEEE must be obtained for all other uses, in any current or future media, including reprinting/republishing this material for advertising or promotional purposes, creating new collective works, for resale or redistribution to servers or lists, or reuse of any copyrighted component of this work in other works. https://doi.org/10.1109/IJCNN48605.2020.9207269
\end{textblock}
}

\usepackage{cite}
\usepackage{amsmath,amssymb,amsfonts}
\usepackage{algorithmic}
\usepackage{graphicx}
\usepackage{textcomp}
\usepackage{xcolor}
\usepackage{booktabs}
\usepackage{threeparttable}
\newtheorem{theorem}{Theorem}
\newtheorem{lemma}{Lemma}
\newtheorem{definition}{Definition}

\usepackage[normalem]{ulem}
\usepackage{verbatim}

\renewcommand\sout{\bgroup\markoverwith
{\textcolor{red}{\rule[.5ex]{2pt}{0.4pt}}}\ULon}

\def\BibTeX{{\rm B\kern-.05em{\sc i\kern-.025em b}\kern-.08em
    T\kern-.1667em\lower.7ex\hbox{E}\kern-.125emX}}

\begin{document}

\title{On the Information Plane of Autoencoders
\thanks{Nicol\'as Tapia acknowledges financial support from the National Agency for Research and Development (ANID) /  Scholarship Program / MAGISTER NACIONAL/2019 - 22191803. The authors acknowledge financial support from ANID-Chile through grant FONDECYT 1171678. Additionally, the authors acknowledge financial support from the Department of Electrical Engineering at Universidad de Chile.}
}

\author{
\IEEEauthorblockN{Nicol\'as I. Tapia}
\IEEEauthorblockA{
Department of Electrical Engineering\\
Universidad de Chile\\
Santiago, Chile \\
nicolas.tapia@ug.uchile.cl}
\and
\IEEEauthorblockN{Pablo A. Est\'evez}
\IEEEauthorblockA{
Department of Electrical Engineering\\
Universidad de Chile\\
Santiago, Chile \\
pestevez@cec.uchile.cl}
}

\maketitle

\copyrightstatement

\begin{abstract}
The training dynamics of hidden layers in deep learning are poorly understood in theory. Recently, the Information Plane (IP) was proposed to analyze them, which is based on the information-theoretic concept of mutual information (MI). The Information Bottleneck (IB) theory predicts that layers maximize relevant information and compress irrelevant information. Due to the limitations in MI estimation from samples, there is an ongoing debate about the properties of the IP for the supervised learning case. In this work, we derive a theoretical convergence for the IP of autoencoders. The theory predicts that ideal autoencoders with a large bottleneck layer size do not compress input information, whereas a small size causes compression only in the encoder layers. For the experiments, we use a Gram-matrix based MI estimator recently proposed in the literature. We propose a new rule to adjust its parameters that compensates scale and dimensionality effects. Using our proposed rule, we obtain experimental IPs closer to the theory. Our theoretical IP for autoencoders could be used as a benchmark to validate new methods to estimate MI in neural networks. In this way, experimental limitations could be recognized and corrected, helping with the ongoing debate on the supervised learning case.
\end{abstract}

\section{Introduction}
\label{sec:previous_ae}

The complexity of deep learning compared with traditional machine learning methods has not allowed a full theoretical understanding of its properties. In particular, it is poorly understood how each hidden layer evolves during training to achieve the end goal of the learning setting. An approach to understand deep neural networks using information-theoretic concepts was first proposed in \cite{tishby2015deep} and further developed in \cite{shwartz2017opening}. The key quantity in this framework is mutual information (MI), derived from the concept of entropy \cite{cover2012elements}. 

The Shannon entropy $H(X)$, or simply entropy, of a discrete random variable (RV) $X\in \mathcal{X}$ with probability mass function $p_X$ is defined as
\begin{equation}
    H(X) = -\sum_{x\in\mathcal{X}}p_X(x)\log (p_X(x)).
    \label{eq:def_entropy_disc}
\end{equation}
Then, the MI between $X$ and another discrete RV $Y\in \mathcal{Y}$ is given by
\begin{equation}
    I(X;Y) = H(X) - H(X|Y) = H(Y) - H(Y|X).
    \label{eq:def_mi_disc}
\end{equation}
When $X$ is a continuous RV with probability density function $f_X$, its differential entropy $h(X)$ is defined as
\begin{equation}
    h(X) = -\int_{x\in\mathcal{X}}f_X(x)\log (f_X(x))dx.
    \label{eq:def_entropy_cont}
\end{equation}
Similarly, the differential MI between $X$ and another continuous RV $Y$ is given by
\begin{equation}
    I(X;Y) = h(X) - h(X|Y) = h(Y) - h(Y|X).
    \label{eq:def_mi_cont}
\end{equation}
The MI measures the dependency between $X$ and $Y$, and attains its minimum, equal to zero, if they are independent. 

Let $A$, $B$ and $C$ form a Markov chain $A\to B\to C$, which means that $C$ is conditionally independent of $A$ given $B$. Then, they satisfy the Data Processing Inequality (DPI) \cite{cover2012elements}:
\begin{equation}
    I(A;B)\ge I(A;C).
    \label{eq:def_dpi}
\end{equation}
Essentially, it means that the information that $B$ contains about $A$ cannot be increased through any transformation of $B$.

Let $X$ and $Y$ be the input and the desired output of a neural network, and let $T$ be an intermediate hidden layer. According to \cite{tishby2015deep}, they form a Markov chain $Y\to X \to T$, satisfying a DPI. The Information Bottleneck (IB) theory predicts that $T$ transforms $X$ so that it maximizes the relevant information about $Y$ while minimizing the information about $X$ \cite{tishby2015deep}. Motivated by this insight, the Information Plane (IP) was proposed to analyze training dynamics.

\begin{definition}[Information Plane]
The Information Plane (IP) is the space with coordinate axes $I(X;T)$ and $I(T;Y)$ at which a hidden layer $T$ in a given training iteration is mapped onto a single point, describing a trajectory during training. 
\label{def:ip}
\end{definition}

According to the IB theory, the learning trajectories of each layer should move towards the point of maximum $I(T;Y)$ and minimum $I(X;T)$. It was experimentally found in \cite{shwartz2017opening} that neural networks exhibit two phases: fitting and compression. The former corresponds to increasing both $I(T;Y)$ and $I(X;T)$, whereas the latter corresponds to decreasing $I(X;T)$ while $I(T;Y)$ increases or stays the same. After compression, each layer stabilizes in a theoretical IB bound. 

The existence of these two phases, and their dependence on the activation functions and the MI estimators have been a topic of ongoing research \cite{saxe2019information, chelombiev2018adaptive, wickstrom2019information}. This motivates the study of classes of neural networks with simpler theoretical behaviors. In this way, an intermediate step can be obtained to recognize undesired experimental limitations and validate new methods to obtain the IP in more general cases.

The IP for a class of neural networks called autoencoders was studied in \cite{yu2019understanding}. An autoencoder outputs a reconstruction $X'$ of its input $X$. It has two components: an encoder and a decoder. The encoder is a first stack of layers that maps $X$ to an encoding $Z$, that is, $Z=\mathrm{Encoder}(X)$. The decoder is a second stack of layers that reconstructs the input from this encoding, that is, $X'=\mathrm{Decoder}(Z)$. Generally, $Z$ has the smallest dimensionality, so the last layer of the encoder is called the \emph{bottleneck layer}.

For simplicity, let assume that both the encoder and the decoder have $L$ layers. Let $\{T_i^E\}_{i=1}^{L-1}$ and $\{T_i^D\}_{i=1}^{L-1}$ be the intermediate layers of the encoder and the decoder, respectively. Then, the autoencoder is represented by
\begin{equation}
    X\to T_1^E \to \cdot\cdot\cdot \to T_{L-1}^E \to Z \to T_1^D \to \cdots \to T_{L-1}^D \to X'.
    \label{eq:ae_sequence}
\end{equation}
Fig.~\ref{fig:ae_model} of Sec.~\ref{sec:result_ae} illustrates this representation for the specific autoencoder used in our experiments.

The IP of Definition~\ref{def:ip} is not suitable for autoencoders as the desired output is the input itself, reducing the plane to a line. It was noted in \cite{yu2019understanding} that an autoencoder  satisfies two DPIs analogous to the supervised learning case: the forward DPI
\begin{equation}
    I(X;T^E_1) \ge \cdots \ge I(X;T^E_{L-1}) \ge I(X;Z),
\end{equation}
and the backward DPI
\begin{equation}
    I(T^D_{L-1}; X') \ge \cdots \ge I(T^D_{1}; X') \ge I(Z; X').
\end{equation}
Both DPIs can be extended to the output and the input layers, respectively. Based on these DPIs, a modified IP was proposed.

\begin{definition}[Information Plane of an Autoencoder] 
The IP of an autoencoder is the space with coordinate axes $I(X;T)$ and $I(T;X')$ at which a hidden layer $T$ is mapped as in Definition \ref{def:ip}. For readability, $I(X;T)$ is called the \emph{input MI} and $I(T;X')$ is called the \emph{output MI} of the layer $T$.
\label{def:ip_ae}
\end{definition}

It was postulated in \cite{yu2019understanding} that the IP curves show two distinct patterns in the form of a bifurcation point depending on whether the bottleneck layer size is larger or smaller than the intrinsic dimensionality of the input data. The authors experimentally found that the IP curves show a compression phase after some critical value for the bottleneck layer size, and that this effect intensifies as the bottleneck gets larger. Our replication of this result is shown in Fig.~\ref{fig:ae_replica} of Sec.~\ref{sec:result_ae}.

Having a compression phase with a large bottleneck means that input information is lost, similar to the supervised learning setting. However, large bottlenecks allow near perfect reconstruction to be achieved. The experimental finding in \cite{yu2019understanding} is conflicting with perfect reconstruction, which requires that all input information is contained at the output layer.

We hypothesize that autoencoders with a large bottleneck layer size do not allow compression. In this work, we theoretically derive the convergence of the IP of autoencoders for different bottleneck layer sizes. Next, we study the limitations of the estimator used in \cite{yu2019understanding} that could have lead to their reported results. The main contributions of this work are the following: a) a theoretical IP of autoencoders with ideal convergence values for the input MI and output MI; and b) an improved adjustment rule for the parameter of the MI estimator used in \cite{yu2019understanding} that allows better agreement between estimations and expected theoretical behaviors.


\section{Theoretical Information Plane of an Autoencoder}
\label{sec:theory}
We assume the common premise that the size $K$ of the bottleneck layer $Z$ restricts the information that can be transferred from the encoder to the decoder. In this section, we derive the theoretical limit of the input MI and the output MI using ideal autoencoders.

\begin{definition}[Ideal Autoencoder] 
An ideal autoencoder minimizes the distance between $X'$ and $X$ (reconstruction error) as much as allowed by its bottleneck layer size $K$.
\label{def:ideal_ae}
\end{definition}

A well-trained autoencoder of enough capacity approximates an ideal autoencoder at the end of training. Therefore, this theoretical limit provides an ideal convergence for each IP curve during training. The specific trajectory followed from initialization to convergence cannot be derived from this analysis because it depends on the optimization process. As a result, we can only provide a sketch of the theoretical IP.

\subsection{Mutual Information Analysis}
\label{sec:mut_info_analysis}

Let $X$ and $T$ be the input and an arbitrary hidden layer of a neural network, respectively. At any given training iteration, the layer $T$ is a deterministic function of $X$. Generally, $X$ has an absolutely continuous component. It was proved in \cite{amjad2019learning} that, in this case, $I(X;T)$ is infinite for almost any selection of weights. In the literature, this problem is avoided either by discretizing $X$ or by measuring MI after adding noise \cite{chelombiev2018adaptive}. In the following, we consider both approaches.

\begin{definition}[Discretization Approach]
Let $A$ and $B$ be two continuous RVs. Let $A_q$ and $B_q$ be their discretized versions obtained using a suitable quantization method. Then, $I(A;B)$ is replaced by $\hat{I}_D(A;B):= I(A_q;B_q)$.
\label{def:mi_discrete}
\end{definition}

\begin{definition}[Noise Addition Approach] 
Let $A$ and $B$ be two continuous RVs. Let $R$ be an independent additive noise. Then, $I(A;B)$ is replaced by $\hat{I}_C(A;B):= I(A;B+R)$.
\label{def:mi_cont}
\end{definition}

\begin{definition}[Unified Approach]
Let $A$ and $B$ be two continuous RVs. Then, $I(A;B)$ is replaced by $\hat{I}(A;B)$, with
\begin{equation}
    \hat{I}(A;B):=\begin{cases}
    \hat{I}_D(A;B) \text{ if discretization is used,}\\
    \hat{I}_C(A;B) \text{ if noise addition is used.}
    \end{cases}
    \label{eq:unif_mi_def}
\end{equation}
\label{def:mi_unif}
\end{definition}

Using $\hat{I}(A;B)$ to analyze autoencoders allows us to prove the following lemma for the output MI.

\begin{lemma}[Output MI] 
$\hat{I}(T;X')=\hat{I}(X;X')$.
\label{lem:output_mi}
\end{lemma}

\begin{IEEEproof}
The discretization approach implies
\begin{equation}
    \hat{I}(T;X')=I(T_q;X'_q)=H(X'_q)-H(X'_q|T_q),
\end{equation}
and the noise addition approach implies
\begin{equation}
    \hat{I}(T;X')=I(T;X'+R)=h(X'+R)-h(X'+R|T).
\end{equation}
Since $X'$ is a deterministic function of any layer $T$, $H(X'_q|T_q)$ equals zero and $h(X'+R|T)$ equals $h(R)$. In both approaches, the output MI is the same for any hidden layer $T$, and equal to $\hat{I}(X;X')$ by taking the particular case $T=X$.
\end{IEEEproof}

Let $\lambda=\lambda(K)$ be an increasing function of the bottleneck layer size $K$ that represents the maximum amount of information that can be transferred from the encoder to the decoder. The following lemma can be proved for the input MI.

\begin{lemma}[Input MI]
An ideal autoencoder satisfies $\hat{I}(X;X')=\hat{I}(X;Z)=\min\{\lambda,\  \hat{I}(X;X)\}$
\label{lem:input_mi}
\end{lemma}

\begin{IEEEproof}
The forward DPI implies that
\begin{equation}
    \hat{I}(X;X)\ge \hat{I}(X;Z)\ge \hat{I}(X;X').
    \label{eq:mini_forward}
\end{equation}
In particular, $X'=X$ achieves the upper bound of $\hat{I}(X;X')$. In this sense, an ideal autoencoder maximizes $\hat{I}(X;X')$ to minimize the reconstruction error. Since the transfer of information is restricted only on the bottleneck layer, the decoder contributes to the maximization of $\hat{I}(X;X')$ by achieving $\hat{I}(X;Z)= \hat{I}(X;X')$ in (\ref{eq:mini_forward}). Additionally, the encoder contributes to the maximization of $\hat{I}(X;X')$ by maximizing $\hat{I}(X;Z)$, which is bounded by $\hat{I}(X;X)$ in (\ref{eq:mini_forward}). Due to the bottleneck restriction, $\hat{I}(X;Z)$ is also bounded by $\lambda$, implying that the achievable maximum is $\min\{\lambda,\ \hat{I}(X;X)\}$.
\end{IEEEproof}

Borrowing the notion from the discrete case, we can interpret $\hat{I}(X;T)$ as the information that layer $T$ preserves from the input $X$. The forward DPI, completed as
\begin{equation}
\begin{split}
 \hat{I}(X;X) \ge \hat{I}(X;T_1^E)\ge\cdots\ge \hat{I}(X;Z) \ge \cdots\\
  \ge \hat{I}(X;T_{L-1}^D) \ge \hat{I}(X;X'),
\end{split}
\label{eq:new_forward_dpi}
\end{equation}
implies that the information is decreased or at most preserved from input to output. For random weights, we expect a significant information loss through the layers. Therefore, when initializing practical autoencoders, we expect a strict inequality in (\ref{eq:new_forward_dpi}) and a small value of $\hat{I}(X;X')$. After training, as stated before, practical autoencoders approximate ideal ones. Lemmas \ref{lem:output_mi} and \ref{lem:input_mi} allow us to prove our main result for ideal autoencoders to complete our characterization of the IP.

\begin{theorem}[Two Patterns] 
Consider an ideal autoencoder with a bottleneck layer size $K$. Let $\lambda=\lambda(K)$ be the maximum amount of information that can be transferred through the bottleneck layer:

\begin{enumerate}
\item[a)] If $\lambda > \hat{I}(X;X)$, then the output MI and the input MI are equal to $\hat{I}(X;X)$ for every hidden layer $T$.

\item[b)] If $\lambda < \hat{I}(X;X)$, then the output MI is equal to $\lambda$ for every hidden layer $T$. Moreover, the encoder has input MIs satisfying
\begin{equation}
\hat{I}(X;X) \ge \hat{I}(X;T_1^E)\ge\cdots \ge \hat{I}(X;Z)=\lambda ,
\label{eq:small_encoder_convergence}
\end{equation}
and the decoder has input MIs satisfying
\begin{equation}
\hat{I}(X;Z) = \hat{I}(X;T_1^D) = \cdots =\hat{I}(X;X')=\lambda .
\label{eq:small_decoder_convergence}
\end{equation}
\end{enumerate}
\label{the:phases}
\end{theorem}

\begin{IEEEproof}
Case a): Suppose $\lambda > \hat{I}(X;X)$. This implies $\hat{I}(X;X')=\hat{I}(X;X)$ according to Lemma~\ref{lem:input_mi}. From Lemma~\ref{lem:output_mi}, this implies $\hat{I}(T;X')=\hat{I}(X;X)$, proving the result for the output MI. On the other hand, $\hat{I}(X;X')=\hat{I}(X;X)$ implies that the forward DPI in (\ref{eq:new_forward_dpi}) achieves equality with $\hat{I}(X;T)=\hat{I}(X;X)$, proving the result for the input MI. 

Case b): Suppose $\lambda < \hat{I}(X;X)$. According to Lemma~\ref{lem:input_mi}, it implies $\hat{I}(X;X')=\hat{I}(X;Z)=\lambda$. From Lemma~\ref{lem:output_mi}, $\hat{I}(X;X')=\lambda$ implies $\hat{I}(T;X')=\lambda$, proving the result for the output MI. In addition, the equality $\hat{I}(X;Z)=\lambda$ and the forward DPI directly imply (\ref{eq:small_encoder_convergence}), proving the result for the input MI in the encoder. Finally, the equality $\hat{I}(X;X')=\hat{I}(X;Z)=\lambda$ implies that the forward DPI in the decoder achieves equality as in (\ref{eq:small_decoder_convergence}), proving the result for the input MI in the decoder.
\end{IEEEproof}

\subsection{Consequences in the Information Plane}

\begin{figure}[tbp]
\centering 
\includegraphics[width=\columnwidth, 
trim={0.5in 0.56in 0.51in 0.4in}, 
clip]{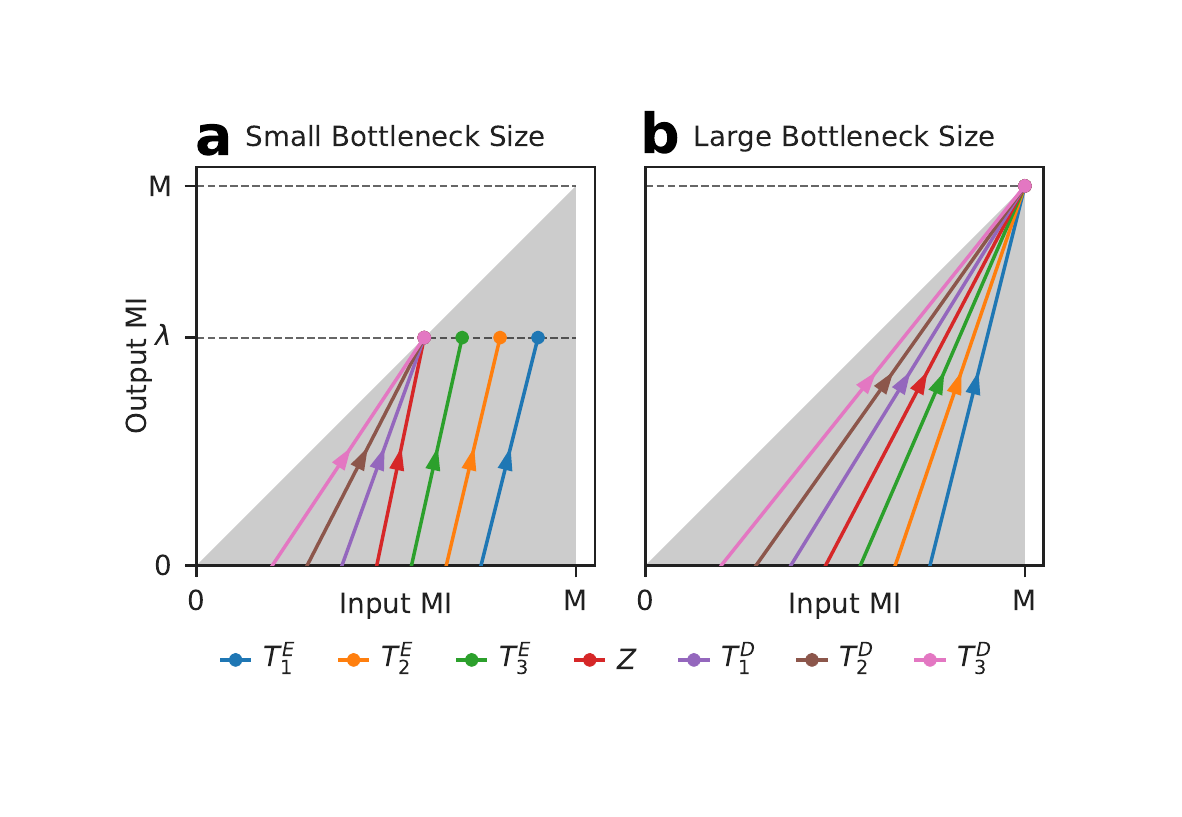}
\caption{Theoretical IP when the bottleneck layer size is (\textbf{a}) small and (\textbf{b}) large. The convergence is highlighted with circles, and the feasible region corresponds to the shaded area. The trajectories to convergence depend on the optimization algorithm, so they were arbitrarily drawn as straight lines.}
\label{fig:sketch}
\end{figure}

For simplicity, we define $M$ as the total information available at the input, i.e., $M=\hat{I}(X;X)$. A direct consequence of the forward DPI in (\ref{eq:new_forward_dpi}) is the feasible region of the IP. Since $ \hat{I}(X;X') \le \hat{I}(X;T)$ and $\hat{I}(T;X') = \hat{I}(X;X')$ (Lemma~\ref{lem:output_mi}), the curves are restricted to the region below the bisector $\hat{I}(X;T)=\hat{I}(T;X')$. Furthermore, $\hat{I}(X;X) \ge \hat{I}(X;T)$, implying that the curves are restricted to the left of the vertical line $\hat{I}(X;T) = M$. In summary, the feasible region is the one contained in the triangle of vertices $(0,0)$, $(M, 0)$ and $(M,M)$, shown as a shaded area in Fig.~\ref{fig:sketch}.

Every layer has the same output MI, equal to $\hat{I}(X;X')$, at each iteration according to Lemma~\ref{lem:output_mi}. Because $\hat{I}(X;X')$ is generally small at initialization, the IP curves will start with input MIs satisfying the inequality of the forward DPI in (\ref{eq:new_forward_dpi}) and with low output MIs, as sketched in Fig.~\ref{fig:sketch}. As the reconstruction error is minimized, $\hat{I}(X;X')$ will grow, implying that the output MIs will grow as well.

Theorem~\ref{the:phases} allows us to derive the convergence of each layer in the IP. For $\lambda > \hat{I}(X;X)=M$, every IP curve will converge to the same input MI and output MI, both equal to $M$. As a result, the IP curves will converge to the point $(M,M)$, at the edge of the feasible region of the IP (see Fig.\ref{fig:sketch}b). On the other hand, for $\lambda < \hat{I}(X;X)=M$, the input MIs of the encoder will converge to a decreasing sequence up to the bottleneck layer were $\hat{I}(X;Z)=\lambda$, and the input MIs of the decoder will converge to the same value $\hat{I}(X;T^D)=\lambda$. The output MI will converge to $\lambda$, so the bottleneck layer and the decoder layers will reach the bisector $\hat{I}(X;T)=\hat{I}(T;X')$, whereas the encoder layers will converge to the interior of the feasible region (see Fig.\ref{fig:sketch}a).

In summary, our theoretical analysis predicts the existence of two distinct patterns in the IP depending on the input information $M$ and the bottleneck layer size $K$ in agreement with \cite{yu2019understanding}. However, we predict neither a bifurcation point nor a compression phase that intensifies will larger $K$. If $K$ is large ($\lambda > M$), all layers converge together on the bisector because they contain all the input information. Therefore, no input information is compressed, which relates to perfect reconstruction. Otherwise, if $K$ is small ($\lambda < M$), some information is compressed through the encoder to achieve the allowed information $\lambda$ at the bottleneck. Then, it is transferred through the decoder preserving as much as possible, without further compression, to minimize the reconstruction error.

The IP of each regime is sketched in Fig.~\ref{fig:sketch} with both the encoder and the decoder having four layers as this is the number of layers used in our experiments. In this sketch, we have drawn the IP curves as straight lines reaching the theoretical convergence. However, these trajectories are arbitrary. The analysis predicts that the output MIs are always equal, but the evolution of the input MIs cannot be deduced from it. Instead, they depend on the optimization algorithm.

The critical value of $K$ that marks the transition between these patterns, also marks the point after which the information is no longer compressed. Therefore, it could be estimated by measuring the MI at the end of training for a range of bottleneck layer sizes. According to \cite{yu2019understanding}, this critical $K$ would approximate the intrinsic dimensionality of the data.

\section{Estimation of Mutual Information}

In practice, the theoretical MI cannot be obtained because the data distribution is unknown, so the MI has to be estimated from samples. Following \cite{yu2019understanding}, we estimate MI during training using the matrix-based estimator proposed in \cite{giraldo2015measures}. It estimates Renyi's $\alpha$-entropy \cite{renyi1961measures}, defined for an RV $X\in\mathcal{X}$ by
\begin{equation}
    H_\alpha(X) = \frac{1}{1-\alpha}\log\int_{\mathcal{X}}f_X^{\alpha}(x)dx.
    \label{eq:renyi_original}
\end{equation}
The standard Shannon entropy is the limit $\alpha\to 1$. In this section, we briefly describe the estimator and our proposed method to adjust its kernel width.

\subsection{Matrix–Based Mutual Information Estimator}
\label{sec:estimator}

Let $X$ be an RV and $x_i\in\mathcal{X}$, $i=1,\ldots,N$ be $N$ realizations of it. Let $\kappa:\mathcal{X}\times \mathcal{X} \to \mathbb{R}$ be an infinitely divisible positive definite kernel that defines a Gram matrix $K\in\mathbb{R}^{N\times N}$ as $K_{ij} = \kappa(x_i,x_j)$. The normalized Gram matrix $A$ is
\begin{equation}
    A_{ij} = \frac{1}{N}\frac{K_{ij}}{\sqrt{K_{ii}K_{jj}}}.
    \label{eq:normalization}
\end{equation}
Let $\lambda_i(A)$ be the $i$-th eigenvalue of $A$. In \cite{giraldo2015measures}, an estimator of the $\alpha$-entropy of $X$ is defined as
\begin{equation}
    S_{\alpha}(A) := \frac{1}{1-\alpha} \log \left( \sum_{i=1}^N \lambda_i(A)^{\alpha} \right).
    \label{eq:estimator}
\end{equation}
Let $Y$ be another RV with normalized Gram matrix $B$. Using the element-wise product $A\circ B$, the joint-entropy estimator is defined as
\begin{equation}
    S_{\alpha}(A, B) := S_\alpha\left( \frac{A\circ B}{\mathrm{tr}(A\circ B)} \right).
    \label{eq:jointestimator}
\end{equation}
From (\ref{eq:estimator}) and (\ref{eq:jointestimator}), an MI estimator is given by:
\begin{equation}
    I_\alpha (A;B) := S_\alpha(A) + S_\alpha (B) - S_\alpha (A, B).
    \label{eq:mitestimator}
\end{equation}
Both $S_\alpha(A)$ and $I_\alpha(A,B)$ are restricted to $[0,\ \log(N)]$.

As in \cite{yu2019understanding}, we set $\alpha=1.01$ to approximate Shannon entropy and choose a gaussian kernel $G_\sigma$ with width $\sigma$, given by
\begin{equation}
    G_\sigma(x_i, x_j) = \beta \exp\left( -\frac{||x_i-x_j||^2}{2\sigma^2} \right),
    \label{eq:gaussian-kernel}
\end{equation}
where $\beta$ is a constant whose value is irrelevant because it is canceled out in the normalized Gram matrix (\ref{eq:normalization}).

\subsection{Kernel Width Selection}
\label{sec:sigma}

The value of the kernel width $\sigma$ is central in the performance of the estimator described in Sec.~\ref{sec:estimator}. The following properties hold for the gaussian kernel:
\begin{align}
    & \lim_{\sigma\to 0} S_\alpha(A) = \log (N),\\
    & \lim_{\sigma\to 0} I_\alpha(A,B) = \log (N),\\
    & \lim_{\sigma\to \infty} S_\alpha(A) =  0,\\
    & \lim_{\sigma\to \infty} I_\alpha(A,B) = 0.
\end{align}
They imply that the value of $\sigma$ controls the operating point of the estimator relative to the bounds because a value too large or too small saturates $S_\alpha(A)$ and $I_\alpha(A;B)$ to 0 and $\log (N)$, respectively. This saturation has to be avoided to have discriminative estimates. Therefore, a suitable value of $\sigma$ has to be determined for an RV $X$ of $d$ dimensions and $N$ samples.

A common rule for the Gaussian kernel is the Silverman's rule of thumb \cite{henderson2012normal} that comes from the literature of density estimation. For the $j$-th dimension of $X$, it is given by
\begin{equation}
    \sigma_j = \left(\frac{4}{2+d}\right)^{1/(4+d)}\hat{\sigma}_j N^{-1/(4+d)},
    \label{eq:silvermann}
\end{equation}
where $\hat{\sigma}_j$ is the empirical standard deviation of the $j$-th dimension. Since $0.92\le\left(4/(2+d)\right)^{1/(4+d)}\le 1.06$, this term can be safely discarded for this application. To study autoencoders in \cite{yu2019understanding}, this rule was further simplified to
\begin{equation}
    \sigma = \gamma N^{-1/(4+d)},
    \label{eq:old-rule}
\end{equation}
where $\gamma > 0$ is an empirically determined constant.

The rule in (\ref{eq:old-rule}) has three main limitations when applied to neural networks. The first one is that an appropriate value of $\gamma$ has to be found experimentally and it can change significantly between variables. This means that a different $\gamma$ could be needed at different layers and even at different iterations as the layers change during training. The second one is that the rule is affected by linear transformations of $X$, whereas Shannon MI is not. Indeed, let $X$ be scaled by $a\in\mathbb{R}$. In (\ref{eq:gaussian-kernel}), this is equivalent to keep the unscaled variable $X$ and to replace $\sigma$ by $\sigma/a$, changing the estimation. This is problematic because neural networks often contain normalization layers such as batch normalization \cite{ioffe2015batch}, and neural layers change their variance during training. In particular, the MI can be overestimated or underestimated depending on whether the variance increases or decreases, respectively. The third limitation is that the rule is affected by dimensionality. To see this, let $X$ have zero mean and unit variance dimension-wise, and let $x_1$ and $x_2$ be two i.i.d. samples. Then, 
\begin{equation}
    \mathbb{E}[||x_1-x_2||^2] = 2d.
    \label{eq:norm_expectation}
\end{equation}
Therefore, the mean square distance is proportional to the number of dimensions. In (\ref{eq:gaussian-kernel}), this means that higher dimensions decrease the effective kernel width on average, increasing the estimated MI value. As a result, neural layers with more units tend to show an overestimated MI. 

The need of improving the adjustment of the kernel width was acknowledged in \cite{wickstrom2019information}. They proposed a method for the supervised learning setting by leveraging the structure induced by the labels at each layer and at each training iteration. This method is not applicable in the general case, and in particular to the case of autoencoders.

\subsection{Proposed Rule for Kernel Width Selection}
\label{sec:sigma_new}

We propose a new rule for the kernel width $\sigma$ that alleviates the aforementioned limitations. Our rule can be understood as augmenting the constant $\gamma$ with variance and dimensionality dependencies. First, we normalize the variable $X$ dimension-wise as
\begin{equation}
    X_j \gets \frac{X_j}{\sqrt{\hat{\sigma}_j^2 + \epsilon}}, \ j=1,\ldots,d,
    \label{eq:norm_scale}
\end{equation}
where $\hat{\sigma}_j$ is the estimated standard deviation for the $j$-th dimension and $\epsilon$ is a small constant to avoid division by zero. This change effectively makes the kernel width different for each dimension and proportional to its standard deviation, returning to the Silverman's rule given by (\ref{eq:silvermann}), so that the kernel is not affected by changes in scale. If $X$ has subsets of components with shared statistics, like channels in an image, then the normalization should be performed by aggregating the statistics of each group, as done in the batch normalization technique. Unlike the Silverman's rule, we additionally modify the rule in (\ref{eq:old-rule}) to
\begin{equation}
    \sigma = \gamma \sqrt{d} N^{-1/(4+d)},
    \label{eq:new-rule}
\end{equation}
so that the kernel width compensates the dimensionality dependency of the mean square distance.

\section{Experiments}
In this section, we present the results of two sets of experiments. First, we validate the proposed kernel width selection rule with a toy problem called \emph{correlated gaussians}. Next, we estimate the IPs of the same autoencoder used in \cite{yu2019understanding} and compare them with the theoretical result of Sec.~\ref{sec:theory}. We tried $\gamma\in [0.1,\  10]$ and selected the value where the dynamics were best shown, although they could be observed in most of them. We used $\epsilon=10^{-8}$ and logarithms of base 2.

\subsection{Toy Problem: Correlated Gaussians}
\label{sec:result_gaussian}

\begin{figure}[tbp]
\centering 
\includegraphics[width=0.65\columnwidth, 
trim={0in 0.15in 0in 0.1in}, 
clip]{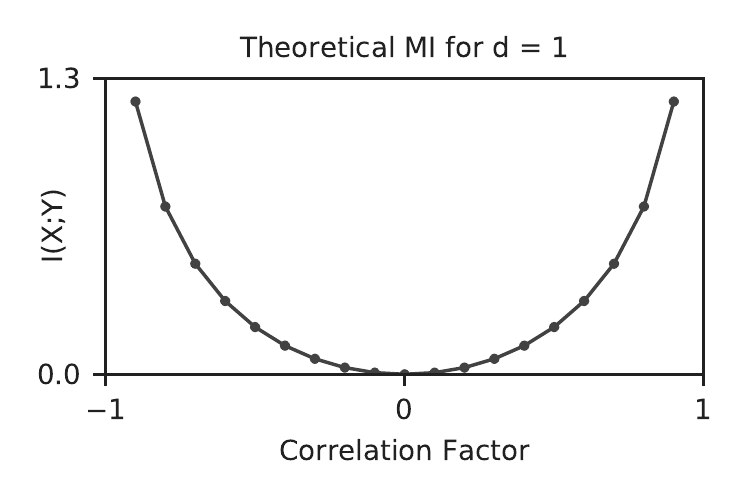}
\caption{Theoretical MI in the \emph{correlated gaussians} problem.}
\label{fig:toy_shannon}
\end{figure}

This problem was used in \cite{belghazi2018mutual} to compare MI estimators, and it is defined as follows. Let $X\sim \mathcal{N}(0,I)$ and $Y\sim \mathcal{N}(0,I)$ be two multivariate normal RVs with $d$ dimensions and with component-wise correlation $\mathrm{corr}(X_i, Y_j) = \delta_{ij}\rho\ \forall i,j\in\{1,\ldots,d\}$, where $\rho\in (-1, 1)$ is the correlation factor and $\delta_{ij}$ is Kronecker’s delta. The problem consists of estimating $I(X;Y)$ from $N$ samples, whose theoretical value is given by
\begin{equation}
    I(X; Y) = - \frac{d}{2}\log(1 - \rho ^2).
    \label{eq:gaussian-mi}
\end{equation}
This theoretical MI is illustrated in Fig.~\ref{fig:toy_shannon} for $d=1$. For other dimensions, it is scaled by $d$ according to (\ref{eq:gaussian-mi}).

\begin{figure}[tbp]
\centering 
\includegraphics[width=0.85\columnwidth, 
trim={0.1in 0.15in 0.1in 0.05in}, 
clip]{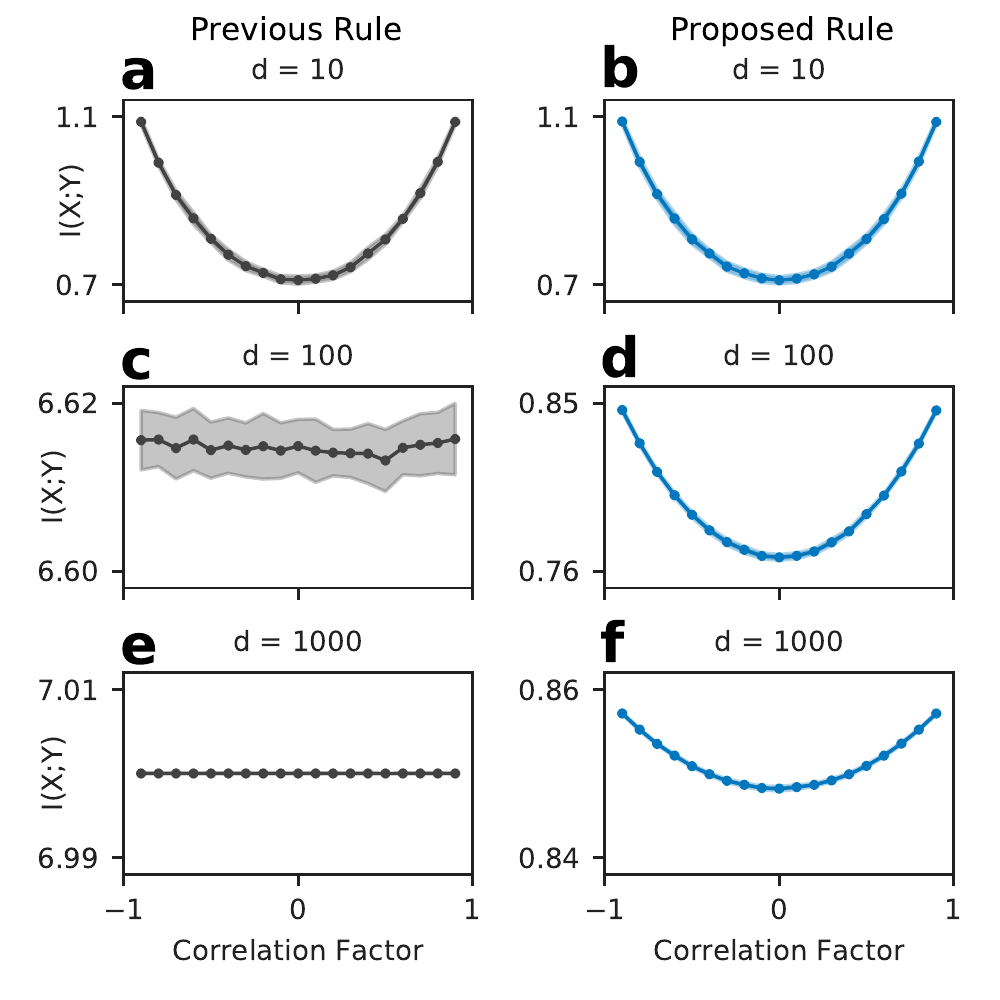}
\caption{Comparison between (\textbf{a},\textbf{c},\textbf{e}) the previous rule and (\textbf{b},\textbf{d},\textbf{f}) the proposed rule for the kernel width selection of the MI estimator using the \emph{correlated gaussians} problem with different dimensions.}
\label{fig:toy_comparison}
\end{figure}

We compare the performance of the MI estimator when selecting the kernel width using the previous rule (\ref{eq:old-rule}) and the rule proposed in Sec.~\ref{sec:sigma_new}. In this problem, the normalization of the variables has no effect because $X$ and $Y$ are standard gaussians. Therefore, the difference lies in whether we add the extra term $\sqrt{d}$ as in (\ref{eq:new-rule}). We evaluate the cases $d\in\{10,\ 100,\ 1000\}$ to cover a range that is commonly found in neural networks, and we set $N=128$ samples. We set $\gamma=2$ for the proposed rule and $\gamma=2\sqrt{10}$ for the rule (\ref{eq:old-rule}), so that they are equivalent when $d=10$. The results are shown in Fig.~\ref{fig:toy_comparison} with the mean and standard deviation of 50 independent runs.

Neither rule can properly approximate the theoretical value. Moreover, the previous rule (\ref{eq:old-rule}) is affected by the saturation effect described in Sec.~\ref{sec:sigma} because the estimations grow close to $\log_2(128)=7$ for $d=100$ and almost exactly for $d=1000$. Even with the proposed rule, the estimations do not grow linearly with $d$ as in the theoretical MI, and the minimum is not zero. However, it can compensate the dimensionality effect to approximate the expected shape of the curve. Therefore, we can expect to approximate tendencies in the IP with this estimator rather than exact values.

\begin{figure}[tbp]
\centering 
\includegraphics[width=0.95\columnwidth, 
trim={0.15in 0.15in 0.15in 0.05in}, 
clip]{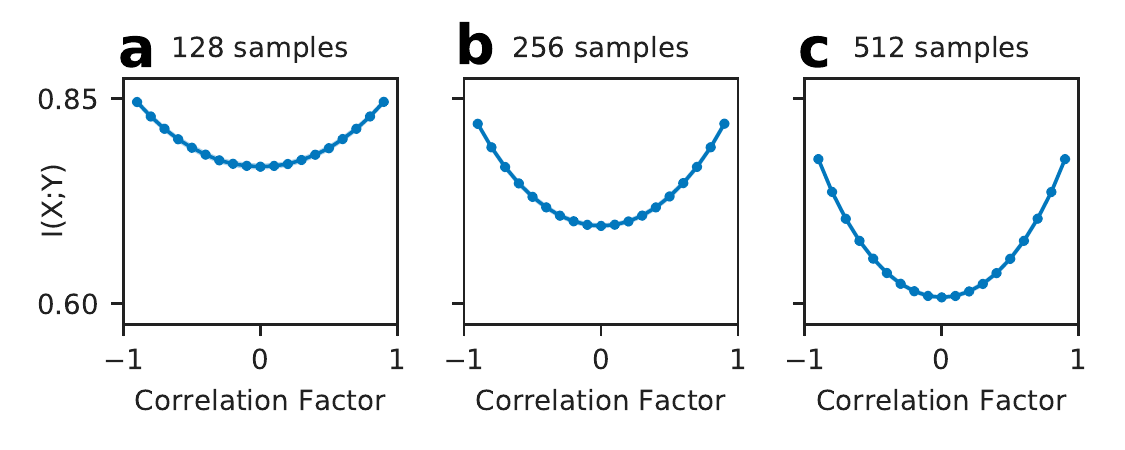}
\caption{Change in the estimated MI when the number of samples is varied, using the \emph{correlated gaussians} problem with 100 dimensions and the proposed rule for the kernel width selection of the MI estimator.}
\label{fig:toy_batch}
\end{figure}

The resolution of the estimation, i.e., the observed range of values, decreases for larger dimensions, but it can be controlled by the number of samples. The estimated MI for $d=100$ using the proposed rule is shown in Fig.~\ref{fig:toy_batch} after increasing the number of samples from 128 to 256 and 512. The resolution progressively increases, improving the confidence in the observed differences. The maximum number of samples is limited by memory constraints because $O(N^2)$ memory is needed to compute the eigenvalues of the Gram matrix.

\subsection{Estimated Information Planes of an Autoencoder}
\label{sec:result_ae}

Following the experiment reported in \cite{yu2019understanding}, we train an autoencoder to reconstruct grayscale images of handwritten digits using MNIST \cite{lecun1998gradient}. This dataset contains 60000 training images and 10000 testing images of $28 \times 28$ pixels. The pixels of each image are normalized to the interval $[0,1]$.

\begin{figure}[tbp]
\centering 
\includegraphics[width=\columnwidth, 
trim={0.1in 0.15in 0.08in 0.2in}, 
clip]{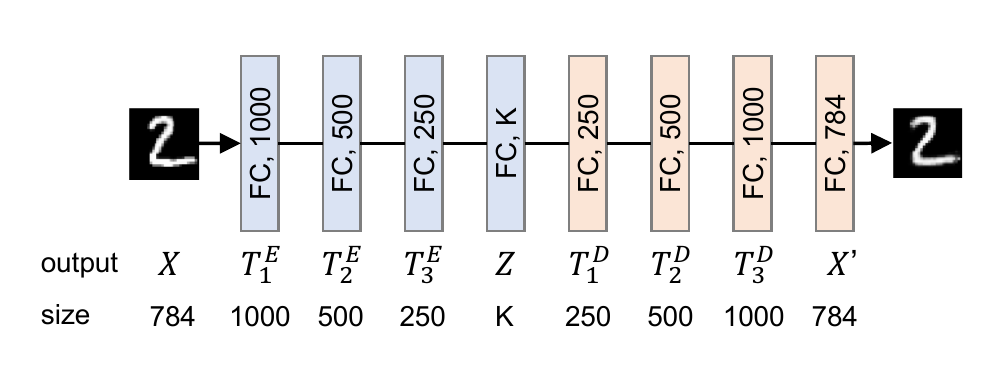}
\caption{Architecture of the autoencoder used in the experiments with fully-connected layers and a variable bottleneck layer size $K$.}
\label{fig:ae_model}
\end{figure}

We use a fully-connected autoencoder, shown in Fig.~\ref{fig:ae_model}, with the same architecture and training process described in \cite{yu2019understanding}. The bottleneck layer size $K$ is varied throughout the experiments. The activation function is sigmoid except for the bottleneck layer where it is linear. The model is trained to minimize the MSE between $X$ and $X'$ using stochastic gradient descent with learning rate 0.1, momentum 0.5, batch size 100, and 100 epochs. To estimate MI at each iteration, we average the results obtained from 10 batches of 512 samples of the testing set. To improve readability, we plot the encoder and the decoder separately in all the experiments. To reduce noise and overplotting in the IPs, we first smooth the estimations by sliding a Hanning window that spans 500 iterations and then we plot a logarithmically spaced subset of iterations.

\begin{figure}[tbp]
\centering 
\includegraphics[width=\columnwidth, 
trim={0.35in 1in 0.3in 0.5in}, 
clip]{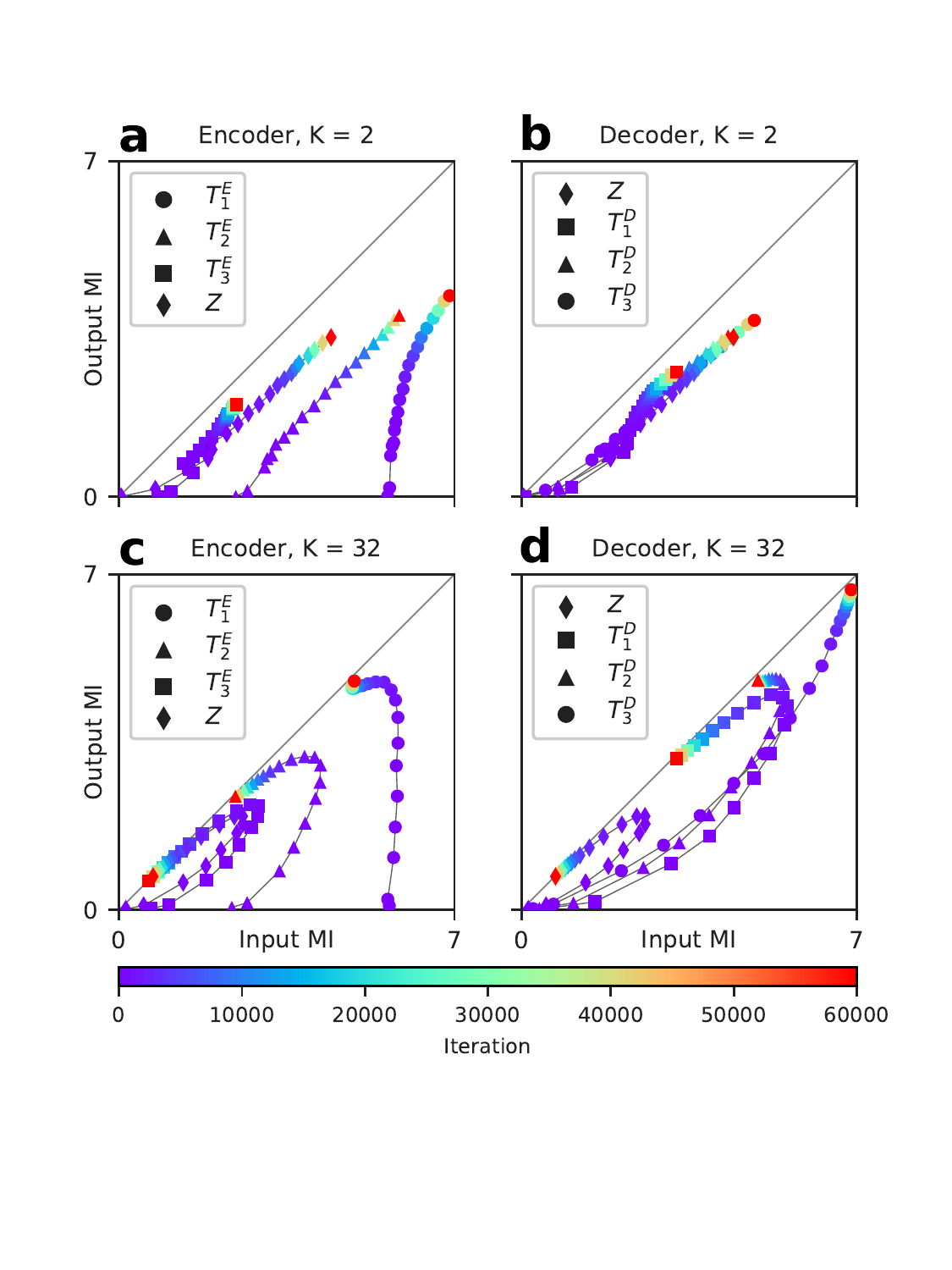}
\caption{Information planes for (\textbf{a},\textbf{b}) $K=2$ and (\textbf{c},\textbf{d}) $K=32$ using the previous rule for the kernel width selection (replication of \cite{yu2019understanding}).}
\label{fig:ae_replica}
\end{figure}

\begin{figure}[tbp]
\centering 
\includegraphics[width=\columnwidth, 
trim={0.35in 1in 0.3in 0.5in}, 
clip]{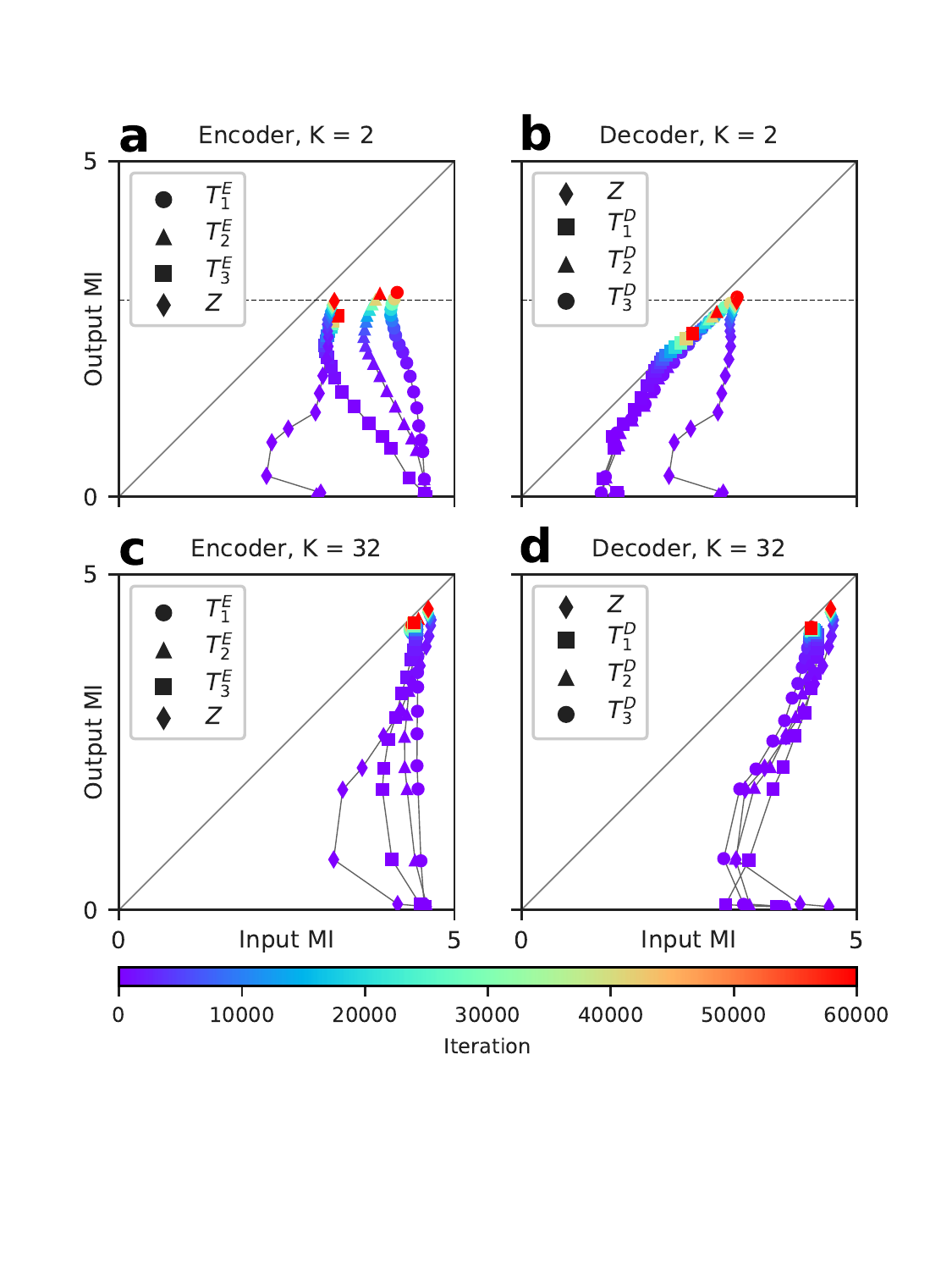}
\caption{Information planes for (\textbf{a},\textbf{b}) $K=2$ and (\textbf{c},\textbf{d}) $K=32$ using the proposed rule for the kernel width selection. The restriction imposed by the bottleneck layer has been highlighted with a horizontal line in the case $K=2$.}
\label{fig:ae_new_both}
\end{figure}

The IPs of the autoencoder for $K=2$ and $K=32$ are computed. We replicate the result of \cite{yu2019understanding} in Fig.~\ref{fig:ae_replica} using the previous rule for the kernel width selection with $\gamma=25$ for the bottleneck layer and $\gamma=5$ for the other layers. For a small $K$ the layers do not get close to the bisector, whereas for a large $K$ the layers show a compression phase towards the bisector. Then, we recompute the IPs in Fig.~\ref{fig:ae_new_both} using our proposed rule with $\gamma=0.8$ for all the layers. In this case, the results follow more closely the theoretical reference shown in Fig.~\ref{fig:sketch}. In particular, there is no compression phase for a large $K$. For $K=2$, there is a visible restriction on the information that can be transferred through the bottleneck layer that results in the compression observed in the sequence of encoder layers.

Comparing the results shown in Figs. \ref{fig:ae_replica} and \ref{fig:ae_new_both} there are two notable differences. The first one is that the layers show different relative magnitudes. For example, in Fig.~\ref{fig:ae_replica}d, the layer $T_3^D$ has a significantly higher input information than $Z$, which is theoretically impossible because it violates the DPI. Conversely, this situation is less significant in the correction shown in Fig.~\ref{fig:ae_new_both}d, probably because the dimensionality effect has been compensated. 

\begin{figure}[tbp]
\centering 
\includegraphics[width=\columnwidth, 
trim={0.1in 0.04in 0.13in 0.03in}, 
clip]{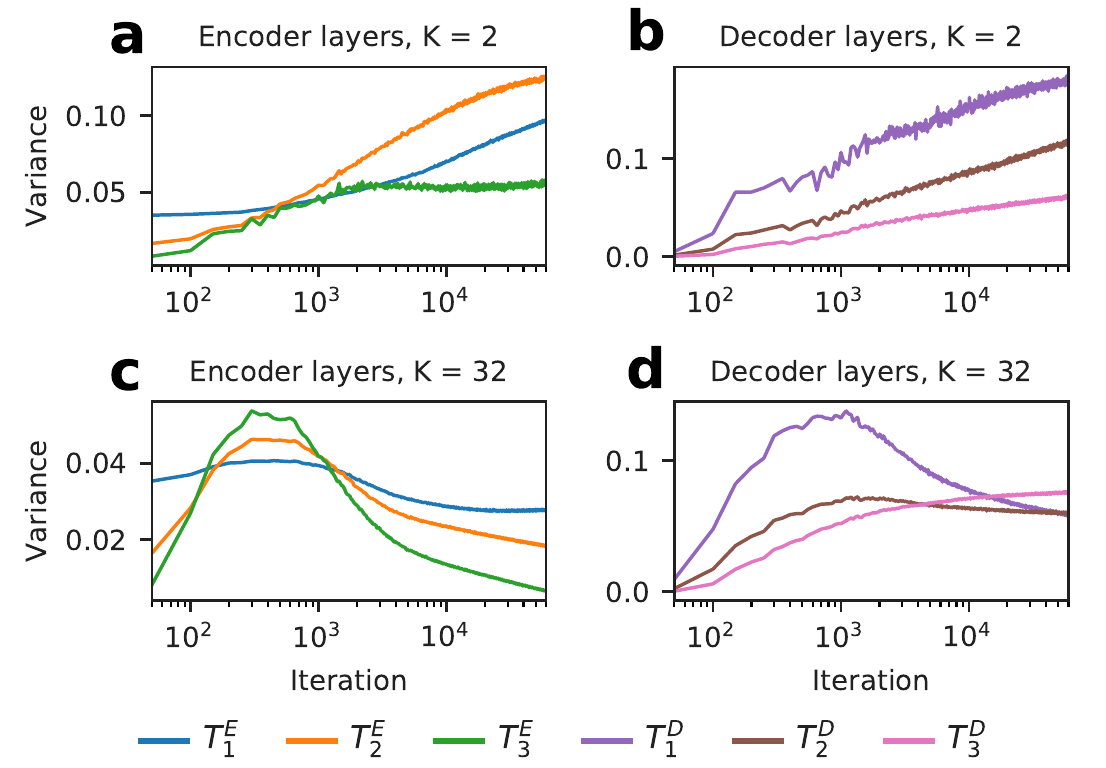}
\caption{Evolution of the average variance of each hidden layer during training for (\textbf{a},\textbf{b}) $K=2$ and (\textbf{c},\textbf{d}) $K=32$.}
\label{fig:ae_variance}
\end{figure}

The second difference is the existence of the compression phase in Figs.~\ref{fig:ae_replica}c-d and the absence of it in Figs.~\ref{fig:ae_new_both}c-d. Based on the theoretical analysis described in Sec.~\ref{sec:sigma}, changes in scale, such as variance, can affect the estimation. A correlation between variance and the estimations was found, which can be observed in the average variance evolution of each layer shown in Fig.~\ref{fig:ae_variance}. For $K=2$, the variance always increases, which correlates with the behavior observed in the IPs of Figs.~\ref{fig:ae_replica}a-b. On the other hand, for $K=32$, the variance starts to decrease after an initial increasing phase, which correlates with the compression phase in Figs.~\ref{fig:ae_replica}c-d. The bottleneck layer variance is not shown in Fig.~\ref{fig:ae_variance} because its variance magnitude is too large compared to the other layers, but the same observations apply. Finally, the variance of $T_3^D$ does not decrease in Fig.~\ref{fig:ae_variance}d, which correlates with the absence of compression for $T_3^D$ in the IP of Fig.~\ref{fig:ae_replica}d. This suggests that the compression phase observed for large $K$ might be caused by variance changes, but an analysis using causal inference would be needed to draw further conclusions.

The DPIs of the autoencoder with the proposed rule are better approximated than in \cite{yu2019understanding}, but they are not fully satisfied yet. For example, the ordering of the layers does not follow strictly the theoretical order even when averaging to compensate for noise in the measurements. Moreover, the theoretical analysis predicts that all layers have the same output MI at every iteration, and this was particularly violated in Fig.~\ref{fig:ae_new_both}b.

\begin{figure}[tbp]
\centering 
\includegraphics[width=\columnwidth, 
trim={0.1in 0.04in 0.13in 0.05in}, 
clip]{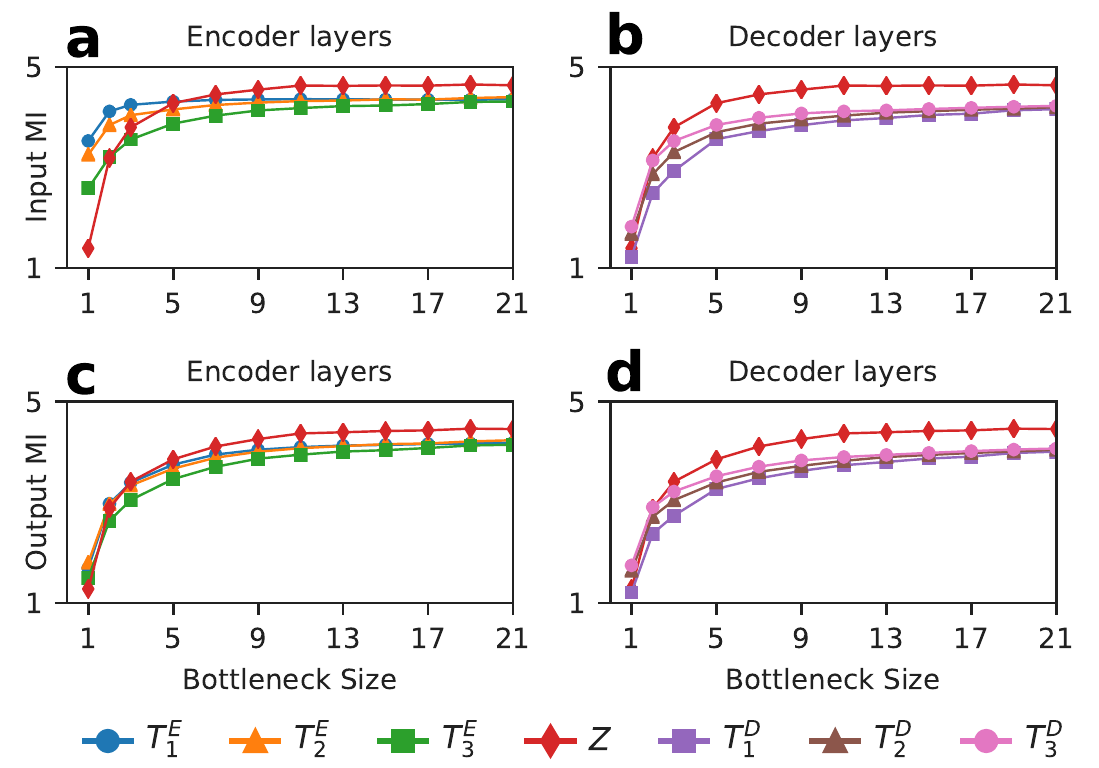}
\caption{Information contained in each hidden layer at the end of training for different bottleneck layer sizes, estimated with the rule proposed in Sec.~\ref{sec:sigma_new}. (\textbf{a},\textbf{b}) MI with the input layer. (\textbf{c},\textbf{d}) MI with the output layer.}
\label{fig:ae_bottleneck}
\end{figure}

Using our proposed kernel width selection rule, we analyze the effect of a range of values of $K$. The input MI and output MI achieved at the end of training by each layer is shown in Fig.~\ref{fig:ae_bottleneck}. Our theoretical prediction is approximated by the experimental results. There are differences in Figs.~\ref{fig:ae_bottleneck}b-d where the MIs should be the same for any $K$, specially for the estimated MI of the bottleneck layer. The MI grows with $K$, which agrees with the premise that larger $K$ allows more information to be transferred. In addition, the result in Fig.~\ref{fig:ae_bottleneck}a approximates the decreasing sequence of the encoder for small $K$, where the input information is decreased from its maximum value to the bottleneck layer value. This sequence shrinks as $K$ increases, except for the bottleneck layer after $K=2$. In Fig.~\ref{fig:ae_bottleneck}a, the compression made by the encoder layers mostly disappears after $K=13$. Beyond this size, all layers are mostly stabilized in their maximum value, where the input MI and the output MI are equal. Hence, $K=13$ is an approximation of the intrinsic dimensionality of the data.

Overall, the estimations using the proposed rule for the kernel width followed reasonably well the expected curves, both in the \emph{correlated gaussians} toy problem and in the information planes of the autoencoder. However, in those cases we had a theoretical reference to compare against the experimental results. We do not know if the estimation errors are small enough to study other problems. As with all estimators, more samples can improve the results, but the memory constraints did not allow to use more than 512 samples at a time. Despite this limitation, the MI estimator used was able to capture expected behaviors for a very high-dimensional setting, which is not the case for other estimators in the literature.

\section{Conclusions}

A particular class of neural networks, the autoencoder, allowed us to obtain theoretical convergences for the IP. They predict that the layers of an autoencoder maximize the information they contain from the input data subject to the restriction imposed by the bottleneck layer size $K$ in the form of a maximum amount of information that can be transferred from the encoder to the decoder. As a result, two patterns appear in the IP depending on whether the bottleneck size is sufficiently large. This is in agreement with what was postulated in \cite{yu2019understanding}, but, contrary to their experimental findings, compression was not observed when $K$ was large enough to allow near perfect reconstruction. Instead, compression was observed for small $K$ and only on the encoder layers, which was linked to the loss of information imposed by the small bottleneck size. To solve this contradiction, we proposed a new rule to adjust the kernel width of the MI estimator used in \cite{yu2019understanding} that compensates for variance effects, as in the original Silverman's rule, and dimensionality effects. This rule allowed us to obtain experimental results that supported our theoretical claims. As future work, these findings have to be further validated using more architectures and datasets.

The absence of information compression was explained by the fact that perfect reconstruction is impossible if any information is lost. However, there exists geometric compression because the number of dimensions is decreased in the bottleneck layer and the dispersion of the variables changes during training as observed in Fig.~\ref{fig:ae_variance}. Because the MI is invariant under bijections, it is inappropriate to measure geometric changes that do not affect the information content. On the contrary, neural networks are sensitive to these transformations, as the ultimate goal for a classification task is to transform the input variable to an output variable that admits a simple linear decision function. These other dimensions of learning that are not captured by the IB theory have been already acknowledged by \cite{amjad2019learning}. It is left as future work to find another measure that captures these other phenomena to complement the theoretical analysis of neural networks.

In agreement with previous works in the IP, estimating MI in neural networks was challenging. We were not able to fully satisfy the theory in the experiments, so more work has to be done in this area. Therefore, our theoretical IP for the autoencoder might serve as a benchmark to assess new approaches to estimate MI in neural networks. In this way, an approach can be validated before using it in the supervised learning scenario where there exists an ongoing discussion on the training dynamics.

\section*{Acknowledgment}
We would like to thank Shujian Yu for providing replication details and Jos\'e Pr\'incipe for useful discussions. Additionally, we would like to thank Rodrigo Carrasco, Jhon Intriago, and Leon Garcia for their idea of summarizing the maximum MI achieved for each bottleneck layer size in a single plot.

\bibliographystyle{IEEEtran}
\bibliography{IEEEabrv, references}

\begin{thebibliography}{10}
\providecommand{\url}[1]{#1}
\csname url@samestyle\endcsname
\providecommand{\newblock}{\relax}
\providecommand{\bibinfo}[2]{#2}
\providecommand{\BIBentrySTDinterwordspacing}{\spaceskip=0pt\relax}
\providecommand{\BIBentryALTinterwordstretchfactor}{4}
\providecommand{\BIBentryALTinterwordspacing}{\spaceskip=\fontdimen2\font plus
\BIBentryALTinterwordstretchfactor\fontdimen3\font minus
  \fontdimen4\font\relax}
\providecommand{\BIBforeignlanguage}[2]{{%
\expandafter\ifx\csname l@#1\endcsname\relax
\typeout{** WARNING: IEEEtran.bst: No hyphenation pattern has been}%
\typeout{** loaded for the language `#1'. Using the pattern for}%
\typeout{** the default language instead.}%
\else
\language=\csname l@#1\endcsname
\fi
#2}}
\providecommand{\BIBdecl}{\relax}
\BIBdecl

\bibitem{tishby2015deep}
N.~Tishby and N.~Zaslavsky, ``Deep learning and the information bottleneck
  principle,'' in \emph{2015 IEEE Information Theory Workshop (ITW)}, 2015, pp.
  1--5.

\bibitem{shwartz2017opening}
R.~Shwartz-Ziv and N.~Tishby, ``Opening the black box of deep neural networks
  via information,'' \emph{Why \& When Deep Learning works: looking inside Deep
  Learning (ICRI-CI paper bundle)}, 2017.

\bibitem{cover2012elements}
T.~Cover and J.~Thomas, \emph{Elements of information theory}, 2nd~ed.\hskip
  1em plus 0.5em minus 0.4em\relax John Wiley \& Sons, 2006.

\bibitem{saxe2019information}
A.~Saxe, Y.~Bansal, J.~Dapello, M.~Advani, A.~Kolchinsky, B.~Tracey, and
  D.~Cox, ``On the information bottleneck theory of deep learning,'' in
  \emph{Int. Conf. Learning Representations (ICLR)}, 2018.

\bibitem{chelombiev2018adaptive}
I.~Chelombiev, C.~Houghton, and C.~O'Donnell, ``Adaptive estimators show
  information compression in deep neural networks,'' in \emph{Int. Conf.
  Learning Representations (ICLR)}, 2019.

\bibitem{wickstrom2019information}
K.~Wickstr{\o}m, S.~L{\o}kse, M.~Kampffmeyer, S.~Yu, J.~Principe, and
  R.~Jenssen, ``Information plane analysis of deep neural networks via
  matrix-based {R}enyi's entropy and tensor kernels,'' \emph{arXiv preprint
  arXiv:1909.11396}, 2019.

\bibitem{yu2019understanding}
S.~Yu and J.~Principe, ``Understanding autoencoders with information theoretic
  concepts,'' \emph{Neural Networks}, vol. 117, pp. 104--123, 2019.

\bibitem{amjad2019learning}
R.~Amjad and B.~Geiger, ``Learning representations for neural network-based
  classification using the information bottleneck principle,'' \emph{{IEEE}
  Trans. Pattern Anal. Mach. Intell.}, 2019, doi: 10.1109/TPAMI.2019.2909031.

\bibitem{giraldo2015measures}
L.~Sanchez, M.~Rao, and J.~Principe, ``Measures of entropy from data using
  infinitely divisible kernels,'' \emph{{IEEE} Trans. Inf. Theory}, vol.~61,
  no.~1, pp. 535--548, 2015.

\bibitem{renyi1961measures}
A.~R{\'e}nyi, ``On measures of entropy and information,'' in \emph{Proc. 4th
  Berkeley Symp. on Math. Statist. and Prob.}, vol.~1, 1961, pp. 547--561.

\bibitem{henderson2012normal}
D.~Henderson and C.~Parmeter, ``Normal reference bandwidths for the general
  order, multivariate kernel density derivative estimator,'' \emph{Statistics
  \& Probability Letters}, vol.~82, no.~12, pp. 2198--2205, 2012.

\bibitem{ioffe2015batch}
S.~Ioffe and C.~Szegedy, ``Batch normalization: Accelerating deep network
  training by reducing internal covariate shift,'' in \emph{Proc. 32nd Int.
  Conf. Machine Learning ({ICML})}, 2015, pp. 448--456.

\bibitem{belghazi2018mutual}
M.~Belghazi, A.~Baratin, S.~Rajeshwar, S.~Ozair, Y.~Bengio, A.~Courville, and
  D.~Hjelm, ``Mutual information neural estimation,'' in \emph{Proc. 35th Int.
  Conf. Machine Learning ({ICML})}, 2018, pp. 531--540.

\bibitem{lecun1998gradient}
Y.~LeCun, L.~Bottou, Y.~Bengio, and P.~Haffner, ``Gradient-based learning
  applied to document recognition,'' \emph{Proc. {IEEE}}, vol.~86, no.~11, pp.
  2278--2324, 1998.

\end{thebibliography}
\end{document}